\PassOptionsToPackage{table}{xcolor}
\documentclass[10pt,logo,copyright]{nvidiatechreport}
\usepackage[numbers,compress]{natbib}

\usepackage{parskip}
\usepackage{xurl}
\usepackage{nicefrac}
\usepackage{array}
\usepackage{multirow}
\usepackage{float}
\usepackage{adjustbox}


\renewcommand{\absfont}{\normalfont\linespread{1.2}\fontsize{10}{12}\selectfont}

\hypersetup{
  pdftitle={VANTAGE-Bench: Evaluating the Infrastructure AI Gap in Vision-Language Models},
  pdfauthor={Zaid Pervaiz Bhat, Nimra Nayyar, Arihant Jain, Lap Fung Chan, John Suchanek, Yu Wang, Varun Praveen, Tomasz Kornuta, Vidya Nariyambut Murali},
}

\title{VANTAGE-Bench: Evaluating the Infrastructure AI Gap in Vision-Language Models}
\newcommand{\headertitle}{VANTAGE-Bench: Evaluating the Infrastructure AI Gap in Vision-Language Models}

\author{Zaid Pervaiz Bhat$^{1,*,\dagger}$, Nimra Nayyar$^{2,*}$, Arihant Jain$^{1,*}$, Lap Fung Chan$^{2,*}$,
John Suchanek$^{2}$, Yu Wang$^{1}$, Varun Praveen$^{1}$, Tomasz Kornuta$^{1}$ and Vidya Nariyambut Murali$^{1}$\\
{\Affilfont $^{1}$NVIDIA \quad $^{2}$Clemson University}}
\authornotes{$^{*}$Equal contribution. $^{\dagger}$Project lead. Correspondence: \texttt{zbhat@nvidia.com}.}

\paperurl{https://vantage-bench.org/}
\begin{document}

\maketitle

\begin{abstract}
  As Vision-Language Models (VLMs) rapidly advance toward physical deployment, the predominant focus has remained on action-oriented Embodied AI evaluated on subject-centric consumer video. This trajectory overlooks a pervasive class of Physical AI: \emph{Infrastructure AI}, which relies on fixed-infrastructure cameras for open-loop insights like safety monitoring and operational logging. How well VLMs process dense, fixed-camera visual data remains insufficiently explored. We introduce \textbf{VANTAGE-Bench}, an evaluation benchmark that measures this ``Infrastructure AI Gap,'' distinguished by four features: 1) \textbf{Domain Relevance}, spanning three operational environments (Logistics, Transportation, and Smart Spaces); 2) \textbf{Modality Breadth}, unifying image and video evaluation to probe semantic, spatial, temporal, and spatio-temporal capabilities; 3) \textbf{Task Format Diversity}, moving beyond multiple-choice to eight task formulations spanning discriminative QA, generative dense captioning, and spatio-temporal grounding; and 4) \textbf{Evaluation Novelty}, a single-pass trajectory-generation protocol for Single Object Tracking and, to our knowledge, the first such evaluation on fixed-camera infrastructure video, scored against specialist trackers. Annotation spans three regimes over 3,346 expert-annotated media assets: video tasks (3,342 annotations over 854 videos), image grounding (4,281 over 1,864 images), and dense detection (27,404 boxes over 628 images).

  Evaluating 17 models zero-shot, we find the shortfall relative to consumer-centric benchmarks is concentrated rather than general. Event verification, referring expressions, and temporal localization fall roughly 9 to 24 points at every model scale, while video question answering stays within 5.3 points of VideoMME and 2D spatial pointing shows no shortfall against BLINK. The temporal pillar is weakest in absolute terms: no system exceeds 55.7 mIoU on temporal localization or 37.3 SODA$_c$ on dense video captioning. On tracking, frontier models come within roughly 5 points of specialist trackers over short horizons but separate as the horizon extends. Open-weight models lead 2D object localization outright, so neither parameter count nor proprietary access accounts for the pattern. The dataset is available at \url{https://huggingface.co/datasets/nvidia/PhysicalAI-VANTAGE-Bench}, and the evaluation harness and public leaderboard at \url{https://vantage-bench.org/}.
\end{abstract}
\abscontent

\section{Introduction}
\label{sec:intro}

As VLMs rapidly advance toward physical deployment, the prevailing research trajectory has heavily indexed on action-oriented Embodied AI and evaluations grounded in subject-centric consumer video (e.g., egocentric recordings, broadcast media). This focus overlooks a pervasive and foundational class of physical AI: \emph{Infrastructure AI}. Relying on fixed-infrastructure cameras (e.g., CCTV, elevated sensors), Infrastructure AI provides open-loop insights for safety monitoring, operational logging, and large-scale spatial reasoning. Because frontier models are predominantly trained on internet-crawled data, they are biased toward standard photographic perspectives. Consequently, their capability to process the dense, oblique, and multi-agent visual data inherent to fixed-camera environments remains insufficiently evaluated, creating a severe implicit domain shift.

Under these conditions, the ``internet-video prior'' fails, and models must rely on pure spatial-temporal logic. A model can achieve state-of-the-art scores on internet video and still perform poorly in the environments where these systems are actually deployed. We term this the \emph{Infrastructure AI Gap}: the discrepancy between a model's performance on general video benchmarks and its reliability as an insight-generating system in dense, fixed-camera infrastructure environments. Closing this gap requires evaluation frameworks designed natively for operational reality. While comprehensive suites like VideoMME \cite{videomme} and MVBench \cite{mvbench} evaluate multiple reasoning dimensions, they ultimately reduce evaluation to a single format: MCQ. MCQ evaluation provides the correct answer as an implicit prior and fails to test if a model can autonomously generate bounding box coordinates or predict explicit temporal boundaries.

To systematically measure and close this capability deficit, we introduce \textbf{VANTAGE-Bench} (\textbf{V}ideo \textbf{An}alysis \textbf{T}asks \textbf{A}cross \textbf{G}eneralized \textbf{E}nvironments). VANTAGE-Bench is the first multi-task benchmark specifically curated for Infrastructure AI, moving beyond single-format MCQ to evaluate eight distinct task formulations across four pillars of operational visual intelligence: Semantic, Spatial, Temporal, and Spatio-Temporal Understanding.

\textbf{Finding:} Our zero-shot evaluation of 17 models shows that the deficit is localized rather than uniform, and the asymmetry is what makes it interpretable. Measured against the same models' published scores on consumer-centric benchmarks, event verification, referring expressions, and temporal localization drop at every scale, while video question answering holds within 5.3 points of VideoMME \cite{videomme} and 2D spatial pointing shows a surplus of 8.3 points against BLINK \cite{blink} at 32B. The same models therefore lose ground on some capabilities and gain it on others within one suite, which places the shortfall in specific capabilities rather than in overall difficulty. Section~\ref{sec:results} tightens this with a prompt- and metric-matched control, and shows that models carrying more physical-AI training data gain 14 to 26 points on tracking and object localization while the two temporal tasks stay nearly untouched --- an asymmetry we trace to the scarcity of densely timestamped video supervision rather than to a general weakness in video understanding. Those two tasks remain the weakest in absolute terms across the entire suite, at frontier scale as much as at 2B.

Our main contributions are:
\begin{enumerate}[leftmargin=*,topsep=2pt,itemsep=1pt]
    \item \textbf{VANTAGE-Bench:} The first multi-task benchmark specifically curated for Infrastructure AI, spanning three deployment domains -- Logistics/Warehouse, Transportation, and Smart Spaces -- addressing the structural distribution shift from egocentric video.
    \item \textbf{Format-Diverse Evaluation:} Eight task formulations spanning discriminative QA, generative dense captioning, and spatio-temporal grounding. Five of the eight require the model to emit bounding boxes, temporal boundaries, or a full trajectory with no candidate set supplied, removing the implicit answer prior that MCQ evaluation provides; only two are multiple-choice and one binary.
    \item \textbf{Tracking Protocol:} A single-pass trajectory-generation protocol for Single Object Tracking (SOT) in VLMs and, to our knowledge, the first such evaluation on fixed-camera infrastructure video spanning open-weight and proprietary models, reported against specialist-tracker, static-box, and random baselines.
    \item \textbf{Expert-Annotated Dataset:} 3,346 media assets across three annotation regimes: 3,342 video-task annotations over 854 videos, 4,281 image-grounding annotations over 1,864 images, and 27,404 detection boxes over 628 images. Annotation density is uneven across regimes by design, and we report the regimes separately rather than as a single aggregate.
    \item \textbf{Open Evaluation Ecosystem:} An open-source evaluation suite and a live public leaderboard with held-out ground truth (Appendix~\ref{app:release}).
    \item \textbf{Empirical Characterization:} A 17-model zero-shot evaluation establishing that the gap is localized rather than uniform --- event verification, referring expressions, and temporal localization fall roughly 9 to 24 points against the same models' published scores, while video question answering and 2D spatial pointing do not --- and that the two temporal tasks hold the lowest ceilings of the eight, unclosed at frontier scale.
\end{enumerate}

\section{Related Work}
\label{sec:related}

\subsection{General Video Benchmarks and the MCQ Bottleneck}
The dominant paradigm in VLM evaluation relies on general-purpose, multi-task benchmarks such as VideoMME \cite{videomme}, MVBench \cite{mvbench}, and EgoSchema \cite{egoschema}. While these comprehensive suites have driven substantial progress in general video understanding, their relevance to operational deployment is limited by two factors. First, they rely on a ``cinematic prior''---utilizing web-sourced, broadcast, or egocentric footage characterized by high-velocity optical flow, human-centric framing, and edited temporal structures. Second, they suffer from the ``MCQ Bottleneck.'' By reducing complex spatio-temporal reasoning to MCQ, these benchmarks provide the correct answer as an implicit prior. This format does not test the open-ended generative capabilities---such as producing precise bounding box coordinates or explicit temporal boundaries---that autonomous Infrastructure AI requires to generate reliable downstream insights.

\subsection{Embodied AI Datasets vs. Classical Surveillance}
The physical AI community has developed extensive datasets for real-world deployment, but these efforts remain fragmented across two divides. First is the operational divide. Datasets like KITTI \cite{kitti}, nuScenes \cite{nuscenes}, and Waymo \cite{waymo} are optimized exclusively for Embodied AI---featuring moving cameras, low-latency requirements, and closed-loop perception. On the other side is classical surveillance, where datasets like UCF-Crime \cite{ucfcrime} and VIRAT \cite{virat} capture the correct fixed-infrastructure perspective, but rely on classical bounding boxes or binary anomaly labels without a natural language interface. Second is the domain divide. Existing enterprise datasets are strictly siloed---focusing solely on either transportation (e.g., AI City Challenge \cite{aicity}), generic crime (UCF-Crime), or robotic manipulation in logistics. VANTAGE-Bench bridges both divides simultaneously. It pairs the fixed-infrastructure perspective of classical surveillance corpora with a natural-language interface, and unifies evaluation across three distinct deployment domains: Transportation, Warehouse, and Smart Spaces. This multi-domain approach ensures models are evaluated on generalized physical reasoning rather than overfitted domain-specific priors.

\subsection{Spatial and Temporal Grounding in Idealized Settings}
Grounding language to spatial and temporal coordinates is a foundational requirement for physical AI, yet existing benchmarks evaluate these capabilities under highly idealized conditions. Spatial grounding and referring expression datasets like RefCOCO \cite{refcoco} and BLINK utilize natural indoor and outdoor photography featuring well-lit, centered, and distinct objects. This fails to evaluate the dense, multi-instance disambiguation required when monitoring infrastructure from an elevated, oblique perspective (e.g., distinguishing between dozens of identical pallets or vehicles). Similarly, temporal grounding benchmarks like Charades-STA \cite{charades} and ActivityNet Captions \cite{activitynet} utilize scripted, continuous human actions where the event of interest dominates the video timeline. In contrast, operational video is characterized by long quiescent periods. VANTAGE-Bench forces models to search through extended periods of inactivity to localize brief, sparse, safety-critical events, which tests temporal attention and contextual recall more directly.

\subsection{Spatio-Temporal Tracking and Grounding}
Tracking objects continuously through dense scenes is among the most demanding spatio-temporal tasks. We position our Single Object Tracking track against four adjacent lines of work.

\paragraph{Classical tracking benchmarks.} LaSOT \cite{lasot}, GOT-10k \cite{got10k}, and MOTChallenge \cite{motchallenge} were designed to evaluate dedicated computer vision pipelines that maintain an explicit internal state (e.g., appearance models, motion priors, and frame-by-frame memory updates). We evaluate four specialist trackers---SAM~3 \cite{sam3}, SAMURAI \cite{samurai}, SUTrack \cite{sutrack}, and MCITrack \cite{mcitrack}---as anchors on our own tracking annotations: they reach approximately 80 Success AUC on the same tracklets and scorer, establishing the level a purpose-built system attains and giving the VLM scores a reference point (Appendix~\ref{app:qualitative}).

\paragraph{Language-conditioned tracking.} TNL2K \cite{tnl2k} evaluates iterative language-conditioned tracking by dedicated architectures, whereas our protocol is single-pass trajectory generation by a general-purpose VLM from a visual first-frame box. SOT removes language in order to isolate visual correspondence over time.

\paragraph{Joint spatio-temporal grounding.} VidSTG \cite{vidstg} and HC-STVG \cite{hcstvg} evaluate joint spatio-temporal grounding, where one score conflates language grounding, spatial localization, and temporal localization. Our pillar design decomposes exactly these: 2D Referring Expressions isolates language to space, Temporal Localization isolates language to time, and SOT removes language to isolate visual correspondence over time.

\paragraph{Tracking with multimodal language models.} Elysium \cite{elysium} and Merlin \cite{merlin} are purpose-trained tracking MLLMs. VideoChat-R1 \cite{videochatr1} and R1-Track \cite{r1track} address zero-shot box-initialized tracking of open VLMs. MMT-Bench \cite{mmtbench} evaluates tracking as multiple-choice box selection, the format limitation our protocol moves past.

\paragraph{Our protocol.} The model is initialized with the target object's bounding box visually overlaid on the first frame. Through a single instruction, it then processes the subsequent multi-frame sequence and outputs a continuous coordinate trajectory, without rolling memory updates. We therefore describe our contribution as a single-pass trajectory-generation tracking protocol for VLMs and, to our knowledge, the first such evaluation on fixed-camera infrastructure video, spanning open-weight and proprietary models.

\begin{figure}[t]
  \centering
  \includegraphics[width=\textwidth]{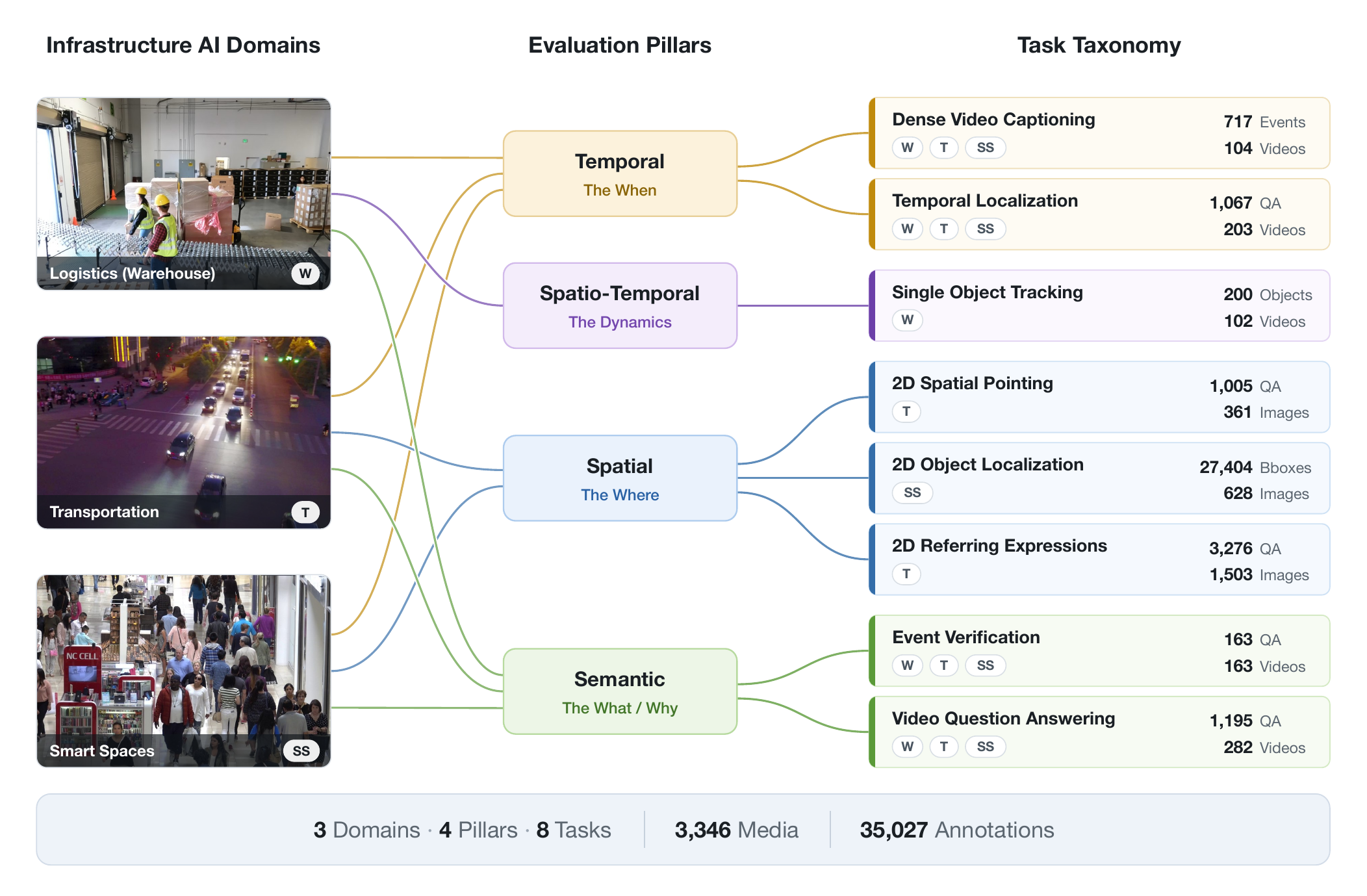}
  \caption{Taxonomy and distribution of VANTAGE-Bench. The benchmark spans three operational domains (W: Warehouse, T: Transportation, and SS: Smart Spaces), evaluating eight distinct tasks across Semantic, Spatial, Temporal, and Spatio-Temporal pillars. In total, the dataset comprises 3,346 media assets across three annotation regimes: 3,342 video-task annotations, 4,281 image-grounding annotations, and 27,404 detection boxes.}
  \label{fig:dataset-overview}
\end{figure}

\section{Dataset and Benchmark Construction}
\label{sec:dataset}

To construct a benchmark capable of testing the Infrastructure AI gap, we curated a multi-modal dataset combining expert-annotated real-world infrastructure footage with a limited set of high-fidelity simulated environments. The dataset is overwhelmingly human-labeled. Automated procedures are used to structure and scale one task format, 2D Spatial Pointing, where the question and its correct answer are both derived deterministically from human-labeled box coordinates (Section~\ref{sec:pointing-engine}). VANTAGE-Bench comprises 3,346 media assets across eight tasks (see Figure \ref{fig:dataset-overview}). The video tasks and 2D Object Localization draw on fixed-infrastructure cameras: static, elevated, wide-angle viewpoints at mounting heights of 8 to 20 feet and viewing angles of 30 to 60 degrees from vertical, including low-light, night, and rain conditions. These are highway ITS and warehouse installations. The two remaining image tracks sample different geometries: 2D Referring Expressions uses oblique aerial views from RefDrone \cite{refdrone}, and 2D Spatial Pointing is derived from object-localization boxes over vehicle-mounted footage. Single Object Tracking uses synthetic elevated-fixed warehouse sequences from PhysicalAI-SmartSpaces \cite{smartspaces}. To support diverse physical reasoning, approximately 20\% of the Video Question Answering and Temporal Localization splits feature high-fidelity synthetic simulations. VANTAGE-Bench is publicly released under the \textbf{NVIDIA Evaluation Data License}, which strictly restricts dataset usage to evaluation and benchmarking purposes, available at \url{https://huggingface.co/datasets/nvidia/PhysicalAI-VANTAGE-Bench}.

\subsection{Design Motivation}
The architecture of VANTAGE-Bench is driven by the disconnect between how VLMs are currently evaluated and how Infrastructure AI operates in production. Existing video benchmarks inadvertently provide models with two strong priors: the ``cinematic prior'' of dynamic, human-centric framing, and the implicit hints embedded within MCQ formats. To evaluate physical grounding, we constructed VANTAGE-Bench around three core design principles:
\begin{itemize}[leftmargin=*, nosep]
    \item \textbf{Ecological Validity:} We strip away internet-video biases by drawing the video tasks and 2D Object Localization from static, elevated, wide-angle infrastructure cameras; the remaining image tracks sample other operational geometries. This forces models to disambiguate dense, multi-instance scenes without the aid of tracking shots or ideal lighting.
    \item \textbf{Format Authenticity:} By moving beyond the MCQ bottleneck, we evaluate autonomous insight generation. VANTAGE-Bench requires models to predict exact temporal boundaries, 2D coordinates, and continuous spatio-temporal trajectories.
    \item \textbf{Dimensional Isolation:} We design specialized task tracks to independently evaluate Semantic, Spatial, Temporal, and Spatio-Temporal intelligence, allowing researchers to isolate the exact locus of model failure.
\end{itemize}

\begin{figure}[t]
  \centering
  \includegraphics[width=\textwidth]{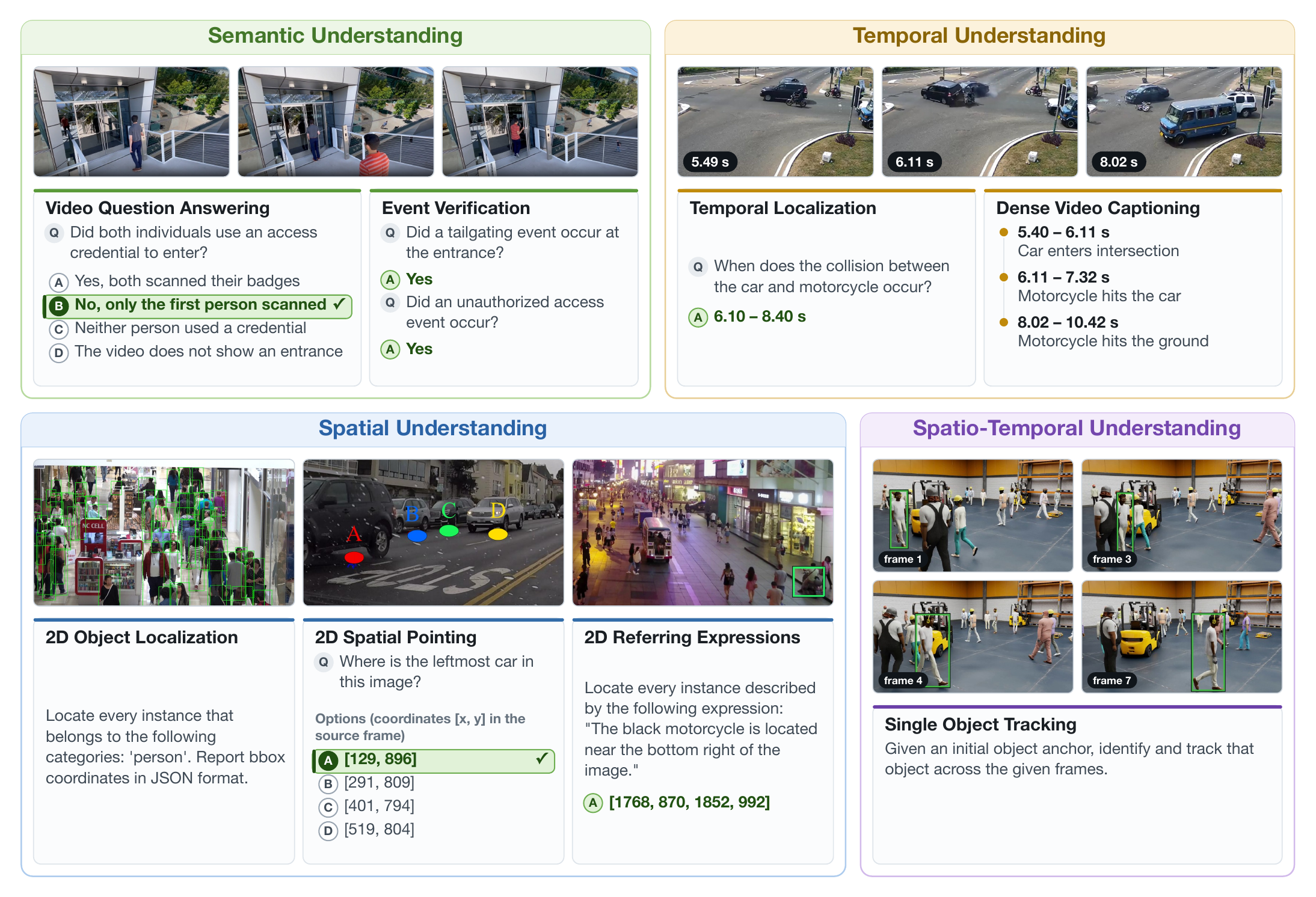}
  \caption{Visual examples of the eight evaluation tasks in VANTAGE-Bench, categorized by their primary reasoning pillar: Semantic, Spatial, Temporal, and Spatio-Temporal Understanding.}
  \label{fig:task-grid}
\end{figure}

\subsection{Task Taxonomy}
\label{sec:taxonomy}
We organize VANTAGE-Bench around four reasoning pillars, evaluating the distinct functional capabilities required for reliable Infrastructure AI. High-level descriptions are provided below, with formal mathematical formulations and evaluation metrics detailed in Appendix~\ref{app:formulations}.

\paragraph{Pillar I: Semantic Understanding.} This pillar evaluates high-level causal logic and operational reasoning via: (1) Event Verification (EV), which requires models to verify operational hypotheses against visual evidence, and (2) Video Question Answering (VQA), which tests multi-step logical reasoning over untrimmed sequences.

\paragraph{Pillar II: Spatial Understanding.} This pillar evaluates precise geometric grounding within infrastructure scenes through: (3) 2D Referring Expressions (RefEx), requiring dense semantic disambiguation to localize targets; (4) 2D Spatial Pointing, evaluating coordinate selection from positional prompts; and (5) 2D Object Localization, establishing a baseline for class-level spatial awareness.

\paragraph{Pillar III: Temporal Understanding.} This pillar focuses on perceiving action duration and sequence boundaries using: (6) Temporal Localization, requiring exact start/end timestamp prediction; and (7) Dense Video Captioning (DVC), a generative task for autonomous localization and description of chronological events.

\paragraph{Pillar IV: Spatio-Temporal Understanding.} This pillar evaluates continuous visual persistence through: (8) Single Object Tracking (SOT). Unlike standard VLM tasks, SOT requires predicting a target's coordinate trajectory across a full sequence, testing the maintenance of spatial context over time.

\subsection{Source Data and Expert Human Annotation}

The core footage for VANTAGE-Bench was curated by NVIDIA to capture hard to source fixed-camera infrastructure perspectives. Real-world footage originates from two U.S. municipalities, San Francisco, CA and Dubuque, IA, supplemented by publicly sourced footage including European footage under license. Consent provenance and the PII obfuscation and verification pipeline are detailed in Appendix~\ref{app:deid}. Annotations were generated by trained professionals---not crowdsourced---using domain-specific rulebooks. Annotators train for months, only the top 10\% of the pool work on benchmark data, and per-task rulebooks are stabilized over up to two months before production. Our quality pipeline uses a multi-tier validation structure: 100\% of initial annotations undergo a second-expert review, with a measured 98\% first-pass acceptance (a 2\% correction rate) and 99.43\% acceptance after revision. A third-tier QA expert then audits a random 10\% subset for consistency across spatial, temporal, and semantic boundaries.

\subsection{Programmatic Task Generation via Pseudo-Labeling}
\label{sec:pointing-engine}

To scale the evaluation of fine-grained spatial reasoning, the Pointing Data Engine converts human-labeled object-localization boxes over vehicle-mounted footage into 2D Spatial Pointing tasks. Both the question and its answer are generated programmatically from KITTI-style labels: the question from a template over relative distances between box coordinates, the correct option deterministically from the same geometry. A VLM is invoked only afterwards, to rewrite the templated question in more natural language; it cannot change which option is correct. Anchoring every pair to human-verified coordinates keeps the answer key tied to expert annotation at scale, and same-class distractors keep the multiple choice difficult.

\subsection{Adaptation of Public and Synthetic Tracks}
To supplement our proprietary real-world data, we selectively adapted highly relevant public and synthetic sources for specific spatial reasoning tasks. For 2D Referring Expressions (RefEx), we integrated images from the RefDrone dataset \cite{refdrone}, leveraging its dense, oblique aerial perspectives to test language-grounded object disambiguation. These queries are human-authored; the programmatic rewriting described in Section~\ref{sec:pointing-engine} applies only to 2D Spatial Pointing. For SOT, we adapted high-fidelity synthetic logistics sequences from the NVIDIA PhysicalAI-SmartSpaces collection \cite{smartspaces}, specifically targeting multi-camera warehouse environments. These sequences were reformatted into our VLM-native, single-pass $[x, y, w, h]$ coordinate prediction protocol. Full details regarding sequence filtering, data reformatting, and temporal sampling densities are provided in Appendix~\ref{app:pipeline}.

\subsection{Safely Simulating Severe Anomalies}
A fundamental challenge in Infrastructure AI is evaluating safety-critical anomalies (e.g., collisions, industrial accidents) that are inherently rare and heavily restricted by privacy regulations. To evaluate these high-stakes scenarios without ethical compromise, approximately 20\% of our VQA and Temporal splits utilize high-fidelity synthetic data generated using NVIDIA Omniverse DRIVE Sim \cite{drivesim}. This ensures VANTAGE-Bench covers severe long-tail physical anomalies that are otherwise absent from purely real-world datasets.

\noindent
\textbf{Benchmark Scope.}
Unlike prior datasets, which are typically restricted to a single modality or reasoning dimension, VANTAGE-Bench jointly evaluates semantic, spatial, temporal, and spatio-temporal reasoning across both image and video domains, and moves past MCQ-only evaluation to generative captioning and coordinate-based grounding. Appendix~\ref{app:data-stats} compares its scale and coverage against representative benchmarks.

\section{Evaluation Framework and Empirical Results}
\label{sec:results}
To quantify the Infrastructure AI capability gap, we evaluate 17 VLMs on VANTAGE-Bench across the four pillars defined in Section~\ref{sec:taxonomy}.

\subsection{Experimental Setup}
\label{sec:setup}
We evaluate 17 models spanning four vendors and 2B to frontier scale. Twelve are open-weight: Qwen3-VL at 2B, 8B, and 32B \cite{qwen3vl}; Qwen3.5 at 9B and 27B \cite{qwen35vl}; Cosmos-Reason2 at 2B, 8B, and 32B \cite{cr2}; Cosmos3 Edge, Nano, and Super \cite{cosmos3}; and Gemma-4-E2B \cite{gemma4}. Five are proprietary: GPT-5.6 Sol \cite{gpt56}, Gemini 3.6 Flash, Gemini 3.5 Flash-Lite, Gemini 3.1 Pro, and Gemini 3.1 Flash-Lite \cite{gemini}. Four of the open-weight families provide a scaling series, which we use in Section~\ref{sec:scaling}. To test out-of-the-box physical reasoning and eliminate fine-tuning biases, all models were evaluated zero-shot. Inference and metric computation (e.g., Story-Oriented Dense Captioning (SODA$_c$) \cite{sodac}, temporal Intersection-over-Union (IoU), spatial trajectory parsing) were conducted using an extended version of the VLMEvalKit harness \cite{vlmevalkit} calibrated for VANTAGE-Bench tasks. Full details regarding our evaluation frameworks, custom metric scripts, and computational setup are provided in Appendix~\ref{app:experimental-details}.

\subsection{Main Results}
We analyze model performance across the four pillars of operational visual intelligence (Table~\ref{tab:main-results}). Performance is uneven across pillars rather than uniformly low. Absolute scores are not directly comparable across tasks, since the metrics differ in what a perfect score demands; Section~\ref{sec:gap} therefore reads each task against an external reference measured the same way.

\begin{table}[htbp]
\centering

\footnotesize
\setlength{\tabcolsep}{3pt}
\begin{tabular}{@{}ll cc ccc cc c c@{}}
\toprule
& & \multicolumn{2}{c}{\textit{Semantic}} & \multicolumn{3}{c}{\textit{Spatial}} & \multicolumn{2}{c}{\textit{Temporal}} & \textit{Sp.-Temp.} & \\
\cmidrule(lr){3-4} \cmidrule(lr){5-7} \cmidrule(lr){8-9} \cmidrule(lr){10-10}
\textbf{Model} & \textbf{Params} & \textbf{VQA} & \textbf{EV} & \textbf{Point.} & \textbf{RefEx} & \textbf{Obj Loc.} & \textbf{Temp Loc.} & \textbf{DVC} & \textbf{SOT} & \textbf{Overall} \\
 & & \tiny(Acc.) & \tiny(M-F1) & \tiny(Acc.) & \tiny(mIoU) & \tiny(F1@.5) & \tiny(mIoU) & \tiny(SODA$_c$) & \tiny(AUC) & \\
\midrule
\multicolumn{11}{@{}l}{\textit{Open-weight models}} \\
Qwen3-VL & 2B & 63.85 & 44.73 & 53.13 & 65.12 & 47.42 & 35.21 & 19.90 & 30.38 & 41.86 \\
Qwen3-VL & 8B & 66.44 & 59.39 & 68.56 & 73.29 & 59.93 & 44.32 & 29.64 & 33.12 & 50.07 \\
Qwen3-VL & 32B & 71.30 & 60.04 & 75.62 & 72.39 & 72.67 & 46.80 & 29.40 & 44.19 & 55.38 \\
Qwen3.5 & 9B & 71.63 & 46.78 & 71.74 & \textbf{77.16} & 76.87 & 42.53 & 27.88 & 23.52 & 48.30 \\
Qwen3.5 & 27B & 67.95 & 55.34 & 75.32 & 76.35 & 85.46 & 36.88 & 26.97 & 24.40 & 49.25 \\
Cosmos-Reason2 & 2B & 64.69 & 55.27 & 59.70 & 56.33 & 73.56 & 38.95 & 28.52 & 26.85 & 45.94 \\
Cosmos-Reason2 & 8B & 67.95 & 64.09 & 68.60 & 70.26 & 83.88 & 47.30 & 32.50 & 37.69 & 54.46 \\
Cosmos-Reason2 & 32B & 70.29 & 73.58 & 74.03 & 50.04 & 4.35$^{\dagger}$ & 50.47 & 30.04 & 33.12 & 47.03$^{\dagger}$ \\
Cosmos3-Edge & 4B & 62.68 & 63.36 & 67.76 & 69.27 & 79.02 & 39.51 & 16.64 & 24.65 & 46.94 \\
Cosmos3-Nano & 16B & 68.95 & 68.88 & 74.83 & 75.57 & 74.11 & 48.04 & 31.42 & 59.21 & 60.67 \\
Cosmos3-Super & 64B & 69.46 & 71.28 & 72.94 & 76.23 & \textbf{86.97} & 51.90 & 29.54 & 64.66 & 63.62 \\
Gemma-4-E2B & 2B & 46.19 & 27.65 & 31.74 & 5.52 & 45.40 & 9.02 & 9.14 & 11.58 & 21.28 \\
\midrule
\multicolumn{11}{@{}l}{\textit{Proprietary models}} \\
Gemini 3.1 Flash-Lite & --- & 68.03 & 63.78 & 63.68 & 52.70 & 71.74 & 37.72 & 32.54 & 46.68 & 52.61 \\
Gemini 3.1 Pro & --- & 71.46 & 68.57 & 72.04 & 67.17 & 77.21 & 45.69 & 35.55 & 67.88 & 62.66 \\
Gemini 3.5 Flash-Lite & --- & 65.61 & 70.41 & 43.08 & 74.46 & 65.73 & 24.00 & 32.29 & 55.77 & 53.25 \\
Gemini 3.6 Flash & --- & 76.82 & \textbf{82.00} & 79.20 & 75.78 & 81.57 & 51.50 & 36.15 & \textbf{75.99} & \textbf{69.52} \\
GPT-5.6 Sol & --- & \textbf{78.08} & 76.14 & \textbf{81.90} & 69.44 & 69.30 & \textbf{55.71} & \textbf{37.28} & 75.02 & 68.04 \\
\bottomrule
\end{tabular}
\caption{VANTAGE-Bench main results. Zero-shot performance of 17 models across the eight tasks, grouped by reasoning pillar. All metrics scaled 0--100; \textit{Overall} is the macro-average of the four pillar scores, and bold denotes the best score per column. Pillar scores are in Table~\ref{tab:pillars} and item-level 95\% bootstrap intervals in Table~\ref{tab:ci}. $^{\dagger}$Output-format regression on Object Localization for this checkpoint; reported unmodified, and it propagates into the aggregates that contain it.}
\label{tab:main-results}
\end{table}

\paragraph{Semantic Understanding.} Semantic understanding is the strongest pillar at the top of the table, though spatial grounding has the higher aggregate across the full model set (Table~\ref{tab:pillars}). Gemini 3.6 Flash leads Event Verification at 82.00 Macro F1 and GPT-5.6 Sol leads Video Question Answering at 78.08, with Cosmos-Reason2-32B and Cosmos3-Super leading the open-weight field on Event Verification at 73.58 and 71.28. Event Verification separates models far more sharply than VQA, spanning 27.65 to 82.00 against 46.19 to 78.08. Smaller models are conservative under uncertainty rather than uniformly weak: Qwen3-VL-8B rejects plausible negatives at a Specificity of 74.58 while reaching only 50.96 in True Event Recall, so it fails to recognize when physical interactions such as collisions or safety violations actually occur. We term this \textbf{causal blindness} and break it down in Appendix~\ref{app:qualitative}; metric definitions are in Appendix~\ref{app:formulations}.

\paragraph{Spatial Understanding.} The spatial pillar is where open-weight models are most competitive. GPT-5.6 Sol leads 2D Spatial Pointing at 81.90, but open-weight models lead the other two spatial tasks outright. Cosmos3-Super leads 2D Object Localization at 86.97, ahead of the best proprietary model at 81.57, and Qwen3.5-9B leads 2D Referring Expressions at 77.16, with Qwen3.5-27B and Cosmos3-Super close behind at 76.35 and 76.23 against 75.78 for the best proprietary model. Single-frame semantic-to-spatial mapping is therefore maturing in open-weight architectures.

\paragraph{Temporal Understanding.} The temporal pillar is the weakest in absolute terms across the entire suite: the best scores on Temporal Localization and Dense Video Captioning, 55.71 mIoU and 37.28 SODA$_c$, sit far below the leader on every other task (Section~\ref{sec:scaling}). This is not confined to open-weight models. On Temporal Localization the best proprietary and best open-weight scores are 55.71 and 51.90, a margin of 3.81 that is the narrowest frontier advantage on any task the frontier leads; on referring expressions and object localization the margin runs the other way, with open-weight models ahead. In Dense Video Captioning, models generate accurate semantic descriptions but fail to localize them in time, which depresses the joint $\text{SODA}_c$ metric; we separate the two components in Appendix~\ref{app:qualitative}.

\paragraph{Spatio-Temporal Understanding.} Single Object Tracking shows the widest separation between proprietary and open-weight models of any task. Gemini 3.6 Flash reaches 75.99 Success AUC and GPT-5.6 Sol 75.02, against 64.66 for the strongest open-weight model, Cosmos3-Super. The spread within the open-weight group is itself large, from 64.66 down to 11.58, and tracks training focus more closely than parameter count. Appendix~\ref{app:tracking-anchors} places these numbers against specialist trackers and shows that the separation grows with the tracking horizon.

\subsection{The Operational Deployment Gap ($\Delta$)}
\label{sec:gap}
To test whether the Infrastructure AI Gap is real, we compare model performance on VANTAGE-Bench against the same models' reported baselines on standard, consumer-focused datasets (Table~\ref{tab:gap-analysis}). We include every model whose developer publishes reference scores: Qwen3-VL reports all five reference benchmarks at 2B, 8B, and 32B, and the Cosmos3 report publishes RefCOCO. Frontier proprietary releases and newer open-weight families increasingly report only high-level multiple-choice cognitive benchmarks and no longer publish zero-shot baselines for granular spatial and temporal grounding, which is itself symptomatic of the evaluation gap we describe.

\paragraph{The deficit is localized, not uniform.} The deltas run in both directions, and that is what makes the pattern informative. Event verification, referring expressions, and temporal localization drop roughly 9 to 24 points across every family and scale we could match. Video question answering sits within 5.3 points, and pointing is flat at the two smaller scales ($-0.7$ and $-0.5$) before turning into a surplus of $+8.3$ at 32B. The same models therefore lose ground on some capabilities and gain it on others within one benchmark, which locates the shortfall in specific capabilities rather than in overall difficulty.

\begin{table}[htbp]
\centering
\small

\begin{tabular}{@{}ll ccc@{}}
\toprule
& & \multicolumn{3}{c}{\textbf{Qwen3-VL}} \\ \cmidrule(lr){3-5}
\textbf{Task} & \textbf{Reference benchmark} & \textbf{2B} & \textbf{8B} & \textbf{32B} \\
\midrule
VQA & VideoMME \cite{videomme} & +2.0 & $-$5.0 & $-$5.3 \\
Event Verification$^{\ast}$ & MLVU \cite{mlvu} & $-$23.6 & $-$18.7 & $-$22.1 \\
2D Spatial Pointing & BLINK \cite{blink} & $-$0.7 & $-$0.5 & +8.3 \\
2D Referring Expressions & RefCOCO \cite{refcoco} & $-$20.5 & $-$15.8 & $-$19.5 \\
Temporal Localization & Charades-STA \cite{charades} & $-$19.3 & $-$11.7 & $-$14.4 \\
\midrule
\multicolumn{5}{@{}l}{\textit{Cosmos3 on RefCOCO \cite{refcoco} (Edge / Nano / Super)}} \\
2D Referring Expressions & RefCOCO \cite{refcoco} & \multicolumn{3}{c}{$-$10.8 \quad / \quad $-$8.7 \quad / \quad $-$13.3} \\
\bottomrule
\end{tabular}
\caption{The Infrastructure AI Gap ($\Delta$ = VANTAGE $-$ reference), for every model whose developer publishes reference scores. Qwen3-VL reports all five reference benchmarks at each scale; the Cosmos3 report publishes RefCOCO only. $^{\ast}$Loosest mapping: Macro F1 against accuracy.}
\label{tab:gap-analysis}
\end{table}

\paragraph{Where the shortfall sits.} Three observations locate it. Models sharing one base diverge sharply by training focus (Section~\ref{sec:scaling}), so the pattern moves with what a model was trained on rather than with the suite as a whole. Proprietary access is no shortcut either: two proprietary models fall below four open-weight models on the overall pillar average (Table~\ref{tab:main-results}). Most directly, the split appears \emph{within} a single task. On dense video captioning, BERTScore-F1 spans 54.2 to 63.9 across models while temporal IoU-F1 spans 33.2 to 39.4 on the same items and the same annotations (Table~\ref{tab:dvc-fluency}): models describe the events correctly and misplace them in time. The failure is in temporal grounding specifically, not in video understanding broadly.

\paragraph{A prompt- and metric-matched control.} Our temporal localization evaluation uses the prompt template the developer published for its own Charades-STA evaluation, so task, metric, and prompt are held constant. Qwen3-VL drops from 56.0 to 44.3 at 8B and from 61.2 to 46.8 at 32B, and the strongest frontier model reaches only 55.7, below what mid-tier open models score on Charades-STA.

\paragraph{A capability with no external baseline.} Model releases claim tracking in their demonstrations, as the Qwen3-VL release does, but none reports a quantitative tracking benchmark, and no prior VLM tracking evaluation targets fixed-camera infrastructure video. Frontier models have therefore been deployed into physical environments without a published baseline for continuous spatio-temporal attention. VANTAGE-Bench supplies one, anchored against specialist trackers in Appendix~\ref{app:tracking-anchors}.

\subsection{Returns to Scale and the Temporal Lag}
\label{sec:scaling}

A prevailing hypothesis is that grounding deficits will resolve through parameter scaling. The four open-weight scaling series let us test this directly (Table~\ref{tab:scaling}).

\begin{table}[htbp]
\centering
\footnotesize
\setlength{\tabcolsep}{2.6pt}

\begin{tabular}{@{}l cccc cc@{}}
\toprule
& \multicolumn{4}{c}{\textit{Open-weight scaling series}} & \multicolumn{2}{c}{\textit{Proprietary}} \\
\cmidrule(lr){2-5} \cmidrule(lr){6-7}
\textbf{Task} & \textbf{Qwen3-VL} & \textbf{Qwen3.5} & \textbf{Cosmos-Reason2} & \textbf{Cosmos3} & \textbf{GPT-5.6} & \textbf{Gemini 3.6} \\
 & \tiny(2B$\rightarrow$32B) & \tiny(9B$\rightarrow$27B) & \tiny(2B$\rightarrow$32B) & \tiny(Edge$\rightarrow$Super) & \tiny Sol & \tiny Flash \\
\midrule
VQA & 63.9$\rightarrow$71.3 (+7.4) & 71.6$\rightarrow$68.0 ($-$3.6) & 64.7$\rightarrow$70.3 (+5.6) & 62.7$\rightarrow$69.5 (+6.8) & \textbf{78.1} & 76.8 \\
Event Verif. & 44.7$\rightarrow$60.0 (+15.3) & 46.8$\rightarrow$55.3 (+8.5) & 55.3$\rightarrow$73.6 (+18.3) & 63.4$\rightarrow$71.3 (+7.9) & 76.1 & \textbf{82.0} \\
Obj. Loc. & 47.4$\rightarrow$72.7 (+25.3) & 76.9$\rightarrow$85.5 (+8.6) & 73.6$\rightarrow$4.4$^{\dagger}$ ($-$69.2) & 79.0$\rightarrow$\textbf{87.0} (+8.0) & 69.3 & 81.6 \\
Ref. Exp. & 65.1$\rightarrow$72.4 (+7.3) & \textbf{77.2}$\rightarrow$76.4 ($-$0.8) & 56.3$\rightarrow$50.0 ($-$6.3) & 69.3$\rightarrow$76.2 (+6.9) & 69.4 & 75.8 \\
Pointing & 53.1$\rightarrow$75.6 (+22.5) & 71.7$\rightarrow$75.3 (+3.6) & 59.7$\rightarrow$74.0 (+14.3) & 67.8$\rightarrow$72.9 (+5.1) & \textbf{81.9} & 79.2 \\
Temp. Loc. & 35.2$\rightarrow$46.8 (+11.6) & 42.5$\rightarrow$36.9 ($-$5.6) & 39.0$\rightarrow$50.5 (+11.5) & 39.5$\rightarrow$51.9 (+12.4) & \textbf{55.7} & 51.5 \\
DVC & 19.9$\rightarrow$29.4 (+9.5) & 27.9$\rightarrow$27.0 ($-$0.9) & 28.5$\rightarrow$30.0 (+1.5) & 16.6$\rightarrow$29.5 (+12.9) & \textbf{37.3} & 36.2 \\
SOT & 30.4$\rightarrow$44.2 (+13.8) & 23.5$\rightarrow$24.4 (+0.9) & 26.9$\rightarrow$33.1 (+6.2) & 24.7$\rightarrow$64.7 (+40.0) & 75.0 & \textbf{76.0} \\
\bottomrule
\end{tabular}
\caption{Returns to scale, smallest to largest member of each open-weight family, with the change in parentheses, alongside the two newest proprietary models. Row-best in bold; markers as in Table~\ref{tab:main-results}.}
\label{tab:scaling}
\end{table}

\paragraph{Returns to scale are task-specific, not pillar-specific.} Scaling helps substantially on some tasks and not at all on others, and the pattern does not align with the four pillars. Temporal Localization improves by an average of 7.5 points from the smallest to the largest family member, so temporal grounding does respond to scale. Object Localization improves with scale in three of the four families; Cosmos-Reason2-32B is the exception, due to an output-format regression rather than a perception result. Qwen3.5 regresses on four of eight tasks despite tripling in size.

\paragraph{The persistent finding is an absolute temporal lag.} Across all 17 models, no system exceeds 55.71 mIoU on Temporal Localization or 37.28 SODA$_c$ on Dense Video Captioning. The same metric on consumer footage runs higher: Qwen3-VL-32B scores 61.2 mIoU on Charades-STA against 46.8 here. Temporal grounding on operational footage is therefore unsolved at present scale, though nothing in our evidence bounds what further scaling might achieve.

\paragraph{Physical-AI training data helps some tasks far more than others.} Qwen3-VL-8B, Cosmos-Reason2-8B, and Cosmos3-Nano share a common base but carry progressively more physical-AI training data. Across that sequence tracking rises 26.1 points, object localization 14.2, and event verification 9.5, while temporal localization moves 3.7 and dense captioning 1.8. All three are evaluated zero-shot. The Cosmos models add physical-AI training over the shared base \cite{cr2, cosmos3}, though the exact composition of that mixture is not public. What the sequence shows is that such training transfers to this suite very unevenly, transforming tracking while barely moving the temporal tasks. Our working hypothesis is a training-mixture asymmetry: spatial-grounding supervision in the form of bounding-box datasets is abundant, while densely timestamped video is scarce. A targeted fine-tuning probe would distinguish a capability gap from an elicitation gap, and we leave this test to future work.

\section{Qualitative Analysis: Failure Modes of Infrastructure AI}
\label{sec:failure-modes}

While Section~\ref{sec:results} quantifies the Infrastructure AI gap, we analyze key failure modes underlying this degradation. Additional quantitative breakdowns are provided in Appendix~\ref{app:qualitative}.

\paragraph{Semantic Illusion (Fluency vs. Grounding).}
Models exhibit strong linguistic fluency but fail to anchor predictions in space and time: they recognise \textit{what} happens without resolving \textit{when} it occurs (Section~\ref{sec:gap}). The effect extends to Event Verification, where smaller models exhibit \textbf{causal blindness}: they correctly reject negatives but fail to detect true events, defaulting to conservative predictions in dense scenes.

\paragraph{Spatial vs. Spatio-Temporal Breakdown.}
Strong single-frame spatial reasoning does not translate to temporal persistence. Weaker models often initialize correctly but fail to maintain object identity as the horizon extends. Two distinct failure modes appear: predictions that stop updating (\textbf{tracking freeze}) and predictions that keep updating but drift off target. Appendix~\ref{app:tracking-anchors} separates these using freeze rate read jointly with Success AUC.

\paragraph{Domain-Specific Vulnerabilities.}
Holding the model and the annotation pipeline fixed, the two pillars degrade in different domains. For Cosmos3-Super, VQA accuracy is comparable in Warehouse and Transportation, at 75.73 and 71.51, and drops sharply only in Smart Spaces, at 55.38, where human activity is least structured and intent is hardest to read from the scene alone. Dense video captioning follows a different order: strongest in Warehouse at 40.61, lower in Smart Spaces at 30.38, and collapsing in Transportation at 8.35. No single domain is uniformly hard. Each pillar has its own worst case, and a model's aggregate score can therefore conceal a domain in which one capability has failed almost completely (Appendix~\ref{app:qualitative}).

\paragraph{Where the Frontier Leads.}
Proprietary access confers no across-the-board advantage; the two newest frontier releases do lead most tasks, but by a fairly uniform margin with tracking the one large outlier. The distinctive frontier capability is temporal persistence rather than spatial grounding, which we quantify in Appendix~\ref{app:frontier}.

\section{Discussion and Limitations}
\label{sec:limitations}

VANTAGE-Bench establishes a baseline for Infrastructure AI evaluation.

\paragraph{Camera geometry.} Our claims are scoped to the camera configurations actually sampled: elevated fixed cameras at 8 to 20 feet and 30 to 60 degrees from vertical for the video tasks and 2D Object Localization, oblique aerial views for 2D Referring Expressions, and vehicle-mounted footage for 2D Spatial Pointing. Real-world fixed-camera deployments vary more widely than this. Eye-level retail and low-angle industrial viewpoints are not covered; the constraint is data rights, since each deployment type requires separate licensing agreements.

\paragraph{Geographic concentration.} Real-world footage comes primarily from two U.S. municipalities, supplemented by publicly sourced and licensed European footage (Section~\ref{sec:dataset}). Coverage is therefore weighted toward North America, and broadening it is future work.

\paragraph{Annotation density.} Counts are uneven across the three annotation regimes because the tasks themselves differ in density. Infrastructure scenes contain many simultaneous objects such as people and vehicles, so a single detection image carries 43.6 boxes on average. The detection task therefore reflects deep labeling by the nature of its definition rather than broad sampling. We report the three regimes separately rather than as one aggregate: video-task annotations, image-grounding annotations, and dense detection boxes.

\paragraph{Quality assurance.} Our acceptance rates come from a sequential second-expert review pipeline, in which a reviewer corrects a prior annotation rather than labeling the item independently. They measure how often expert work passes review, not agreement between independent annotators.

\paragraph{Programmatic question generation.} For 2D Spatial Pointing, the correct answer is fixed by human-labeled geometry and cannot be altered by the rewriting stage. The rewritten question text is nonetheless not constrained to preserve the uniqueness of its spatial cue in dense scenes, and we do not currently verify that it does. A targeted uniqueness audit of the rewritten questions is planned.

\paragraph{Tracking is synthetic only.} The Single Object Tracking track is built from synthetic warehouse sequences, so its conclusions are limited to that distribution and do not yet separate tracking failure from synthetic-domain mismatch. The specialist-tracker and horizon analyses in Appendix~\ref{app:tracking-anchors} establish that the task is well posed, but they do not substitute for real-data validation. Real tracking sequences are planned for a future release.

\paragraph{Model coverage and reported uncertainty.} Our evaluation is strictly zero-shot; domain-specific fine-tuning is left to future work. Extreme weather and lighting conditions are not systematically covered. Inference settings were tuned per model, following each developer's published guidance where it exists and our own prompt-sensitivity checks otherwise; Appendix~\ref{app:inference} reports the configuration each model was finally evaluated under rather than the alternatives we tried. The intervals we report (Appendix~\ref{app:bootstrap}) therefore cover item sampling only, and do not capture how far a score would move under a different prompt or decoding configuration.

\paragraph{Planned extensions.} Coverage is uneven across tasks and domains: 2D Spatial Pointing and 2D Referring Expressions are currently drawn from a single domain, and the tracking track is synthetic. We are extending every task with additional real-world footage spanning all three deployment domains rather than a subset, and we are broadening the camera configurations sampled within each. We also plan a human performance baseline on a representative subset of each task, which would calibrate task difficulty independently of annotation quality. Beyond the current suite, we plan to add task tracks that operational deployments require and this release does not cover, including multi-object tracking and 3D grounding.

\paragraph{Ethical considerations.} The automation of fixed-camera monitoring carries privacy implications. All raw assets underwent a de-identification pipeline combining automated obfuscation with human verification, documented in Appendix~\ref{app:deidentification}. VANTAGE-Bench is released under a license prohibiting biometric identification and demographic profiling. We emphasize that the benchmark evaluates physical and causal reasoning rather than identity recognition, and we advocate for human-in-the-loop oversight in downstream operational deployments.

\section{Conclusion}
\label{sec:conclusion}

We introduced VANTAGE-Bench, an evaluation suite for VLMs in open-loop Infrastructure AI settings. Moving beyond the priors of internet video and single-format MCQ evaluation, we built an 8-task taxonomy spanning semantic, spatial, temporal, and spatio-temporal reasoning over 3,346 expert-annotated media assets, and evaluated 17 models zero-shot.

The resulting picture is uneven rather than uniformly poor. Open-weight models lead 2D Object Localization outright, and they edge ahead on 2D Referring Expressions by a margin comparable to its confidence interval. The deficit against consumer-centric benchmarks concentrates in event verification, referring expressions, and temporal grounding, while pointing is flat at smaller scales and turns into a surplus at 32B. Returns to scale are task-specific rather than pillar-specific. What persists across all 17 models is an absolute lag on the two temporal tasks, which no system has yet closed, and a widening separation on tracking as the horizon extends. We hope VANTAGE-Bench provides a useful framework for evaluating physical reasoning in operational settings, and that its open leaderboard makes progress on these specific deficits measurable.

\section*{Acknowledgements}
\looseness=-1
We thank Paris Zhang, Yilin Zhao and Zheng Liu for developing portions of several task tracks, including Event Verification; Chintan Shah, Sumeeth Nagaraja and Ratnesh Kumar for data sourcing, discussions and evaluation feedback; and Yao Xu and the NVIDIA Data Factory team for video sourcing and human annotation. Tsung-Yi Lin, Ke Ding and Ming-Yu Liu provided evaluation infrastructure and model access. We thank Professor Carrie Russell of Clemson University for collaboration throughout, and Ahmet Dokmeci for the leaderboard evaluation pipeline. This research used in part resources on the Palmetto 2 cluster at Clemson University under National Science Foundation awards MRI 1228312, II NEW 1405767, MRI 1725573, and MRI 2018069. The views expressed in this article do not necessarily represent the views of NSF or the United States government.

\clearpage
\appendix

\section{Data Statistics and Benchmark Comparison}
\label{app:data-stats}

This appendix places VANTAGE-Bench alongside the benchmarks it is most often compared against and records how its annotations are distributed. Table~\ref{tab:bench-stats} reports scale, modality, and pillar coverage for each.

\begin{table}[!ht]
\centering
\small
\renewcommand{\arraystretch}{1.15}

\setlength{\tabcolsep}{5pt}
\adjustbox{max width=\textwidth}{%
\begin{tabular}{@{}l l r r l l@{}}
\toprule
\textbf{Benchmark} & \textbf{Modality} & \textbf{\# Media} & \textbf{\# Annot.} & \textbf{Pillars} & \textbf{Annotation Source} \\
\midrule
VideoMME & Video & 900 & 2,700 & Semantic & Human \\
BLINK & Image & 3,683 & 1,906 & Spatial & Human, Existing \\
RefCOCO\_avg & Image & 3,982 & 30,969 & Spatial & Human, Existing \\
ODinW13 \cite{odinw} & Image & 4,608 & 10,966 & Spatial & Human, Existing \\
CharadesSTA & Video & 1,334 & 3,720 & Temporal & Human, PL \\
ActivityNet Cap. & Video & 5,044 & 17,750 & Temporal & Human \\
\midrule
\textbf{VANTAGE-Bench} & \textbf{Image + Video} & \textbf{3,346} & \textbf{35,027}$^{\dagger}$ & \textbf{All four} & Human, Simulated, PL, Existing \\
\bottomrule
\end{tabular}%
}
\caption{Cross-benchmark comparison. All statistics reflect available evaluation test sets. \textit{Pillars} is the reasoning taxonomy each benchmark covers. $^{\dagger}$Shown as a single total for comparability with the other rows; we report our annotations by regime elsewhere.}
\label{tab:bench-stats}
\end{table}

Each of the three annotation regimes listed in Section~\ref{sec:dataset} is comparable to or larger than its established counterpart in Table~\ref{tab:bench-stats}: the video tasks carry more annotations than VideoMME (3,342 against 2,700), though over slightly fewer videos (854 against 900), and image grounding exceeds BLINK's test set (4,281 against 1,906). Tracking is the one smaller track. Detection dominates the aggregate because infrastructure scenes are dense, at 43.6 boxes per image against ODinW-13's 2.4, which reflects scene density rather than broad sampling.

\section{Formal Task Formulations and Evaluation Metrics}
\label{app:formulations}

In this section, we provide the rigorous definitions for the mappings and evaluation protocols used in VANTAGE-Bench.

\subsection{Notation and Definitions}
Each task is stated in Table~\ref{tab:formulations} as a function from inputs to a structured output, with its primary metric. We define the following input and output spaces to standardize the task formulations:
\begin{itemize}[leftmargin=*, nosep]
    \item \textbf{Inputs:} $\mathcal{I}$ (Image); $\mathcal{V}$ (Video); $\mathcal{Q}$ (Query); $B_{t=0}$ (Initial Box).
    \item \textbf{Outputs:} $\tau$ (Temporal segment $[t_{start}, t_{end}]$); $C$ (Caption); $a$ (Selection from candidates); $\{B\}$ (Set of coordinates).
\end{itemize}

\begin{table}[htbp]
\centering
\small
\renewcommand{\arraystretch}{1.15}

\begin{tabularx}{\textwidth}{@{}l l X l l@{}}
\toprule
\textbf{Pillar} & \textbf{Task} & \textbf{Mathematical Formulation} & \textbf{Primary Metric} & \textbf{Additional} \\
\midrule
\multirow{2}{*}{Semantic} & EV & $f(\mathcal{V}, \mathcal{Q}_{hyp}) \rightarrow \{0, 1\}$ & Macro F1 & Sens., Spec. \\
 & VQA & $f(\mathcal{V}, \mathcal{Q}, \{A,B,C,D\}) \rightarrow a \in \{A,B,C,D\}$ & Accuracy & --- \\
\midrule
\multirow{3}{*}{Spatial} & RefEx & $f(\mathcal{I}, \mathcal{Q}) \rightarrow [x, y, w, h]$ & mIoU & Precision@0.5 \\
 & Pointing & $f(\mathcal{I}, \mathcal{Q}, \{A,B,C,D\}) \rightarrow a \in \{A,B,C,D\}$ & Accuracy & --- \\
 & Obj Loc. & $f(\mathcal{I}, \mathcal{Q}_c) \rightarrow \{B_k\}_{k=1}^M$ & F1@0.5 & COCO mAP \\
\midrule
\multirow{2}{*}{Temporal} & Temp Loc. & $f(\mathcal{V}, \mathcal{Q}) \rightarrow [t_{start}, t_{end}]$ & mIoU & Recall@0.5 \\
 & DVC & $f(\mathcal{V}) \rightarrow \{(\tau_i, C_i)\}_{i=1}^N$ & SODA$_c$ & BERTScore-F1, IoU-F1 \\
\midrule
Spatio-Temp. & SOT & $f(\mathcal{V}, B_{t=0}) \rightarrow \{B_t\}_{t=1}^T$ & Success AUC & Mean IoU \\
\bottomrule
\end{tabularx}
\caption{Formal task definitions for VANTAGE-Bench. The primary metric is the one reported in Table~\ref{tab:main-results} and on the public leaderboard.}
\label{tab:formulations}
\end{table}

\subsection{Metric Clarifications}
\begin{itemize}[leftmargin=*, nosep]
    \item \textbf{SODA$_c$:} Evaluates narrative quality by matching predicted event sequences to ground truth based on temporal overlap and BERTScore-based \cite{bertscore} linguistic similarity.
    \item \textbf{Success AUC:} For SOT, the Success Plot counts frames where predicted IoU exceeds threshold $\sigma$, calculated by varying $\sigma \in [0,1]$.
    \item \textbf{Macro F1:} Penalizes models that exploit majority-class bias in binary Event Verification.
\end{itemize}

\section{Pipeline Details and Data Adaptation}
\label{app:pipeline}

\subsection{Data Adaptation for SOT and RefEx}
\textbf{2D Referring Expressions:} We utilized RefDrone because its elevated perspective mimics infrastructure cameras. We ported prompts requiring dense disambiguation between identically colored vehicles based on relative spatial relationships.

\textbf{Single Object Tracking:} Adapted from the MTMC 2025 subset of PhysicalAI-SmartSpaces. We filtered 17 warehouse scenes featuring workers and robotic profiles (e.g., Nova Carter). Sequences include an initial ``visual anchor'' followed by a sequence of frames sampled at densities of 8, 16, and 32 frames.

\section{Extended Analysis: Deconstructing the Infrastructure AI Gap}
\label{app:qualitative}

This section provides detailed quantitative and qualitative analysis of the failure modes summarized in Section~\ref{sec:failure-modes}. While Section~\ref{sec:results} establishes the magnitude of the Infrastructure AI gap, the analyses below isolate the underlying mechanisms responsible for model failure.

\subsection{Semantic Illusion: Fluency vs. Grounding}

We observe a systemic disconnect between linguistic fluency and physical grounding, which we term the \textbf{Semantic Illusion}.

\paragraph{Dense Video Captioning.}
Models generate semantically coherent descriptions (high BERTScore-F1) but fail to temporally localize events, resulting in low joint performance ($\text{SODA}_c$). This indicates that models capture \textit{what} happens but fail to resolve \textit{when} it occurs, leading to temporal smearing across events. Table~\ref{tab:dvc-fluency} separates the two components.

\begin{table}[htbp]
\centering
\small

\begin{tabular}{@{}lccc@{}}
\toprule
\textbf{Model} & \textbf{Semantic Fluency} & \textbf{Discrete Loc.} & \textbf{Joint Metric} \\
& (BERTScore-F1) & (IoU-F1) & (SODA$_c$) \\ \midrule
Gemini 3.1 Pro & \textbf{63.86} & \textbf{39.44} & \textbf{35.55} \\
Cosmos-Reason2-8B & 58.08 & 36.70 & 32.50 \\
Qwen3-VL-8B & 58.66 & 33.24 & 29.64 \\
Cosmos-Reason2-32B & 54.74 & 33.95 & 30.04 \\
Cosmos3-Super & 54.24 & 33.39 & 29.54 \\
\bottomrule
\end{tabular}
\caption{Dense Video Captioning decomposed into its semantic and temporal components, with the joint metric for reference. Scaled 0--100.}
\label{tab:dvc-fluency}
\end{table}

\paragraph{Event Verification: Causal Blindness.}
The same disconnect appears in Event Verification, though the direction of the imbalance depends on the model rather than on scale alone (Table~\ref{tab:ev-breakdown}). Qwen3-VL-8B fails to detect true events while maintaining strong specificity, at 50.96 against 74.58, a bias toward conservative predictions under uncertainty rather than an inability to reason about the scene. Cosmos-Reason2-8B shows the opposite profile, at 71.15 sensitivity against 57.63 specificity, accepting events that did not occur. Macro F1 conceals both, which is why we report the two components separately.

\begin{table}[htbp]
\centering
\small

\begin{tabular}{@{}lccc@{}}
\toprule
\textbf{Model} & \textbf{Macro F1} & \textbf{True Event Recall} & \textbf{True Negative Recall} \\
& & (Sensitivity) & (Specificity) \\ \midrule
Gemini 3.6 Flash & \textbf{82.00} & \textbf{87.50} & \textbf{76.27} \\
GPT-5.6 Sol & 76.14 & 77.88 & \textbf{76.27} \\
Cosmos-Reason2-32B & 73.58 & 83.65 & 62.71 \\
Cosmos3-Super & 71.28 & 72.12 & 72.88 \\
Gemini 3.1 Pro & 68.57 & 71.15 & 67.80 \\
Cosmos-Reason2-8B & 64.09 & 71.15 & 57.63 \\
Qwen3-VL-8B & 59.39 & 50.96 & 74.58 \\
\bottomrule
\end{tabular}
\caption{Event Verification causal breakdown ($n=163$; 104 positive, 59 negative). Sensitivity and specificity are reported separately, since Macro F1 alone conceals the asymmetry.}
\label{tab:ev-breakdown}
\end{table}

\subsection{Spatial vs. Spatio-Temporal Breakdown}

We identify a key architectural limitation: \textit{static spatial grounding does not imply temporal persistence}. Table~\ref{tab:spatial-gap} contrasts the two for six representative models.

The dissociation is sharpest in the open-weight group. Qwen3.5-27B reaches 85.46 on 2D Object Localization and 76.35 on 2D Referring Expressions, among the strongest spatial results in the suite, yet scores 24.40 on tracking. Cosmos3-Super posts comparable spatial numbers and reaches 64.66. Spatial grounding is therefore not sufficient for temporal persistence, and the two are not acquired together.

\begin{table}[htbp]
\centering
\small

\begin{tabular}{@{}lcccc@{}}
\toprule
\multirow{2}{*}{\textbf{Model}} & \multicolumn{3}{c}{\textit{Spatial}} & \textit{Spatio-Temporal} \\ \cmidrule(lr){2-4} \cmidrule(lr){5-5}
& \textbf{Pointing} & \textbf{RefEx} & \textbf{Object Loc.} & \textbf{SOT} \\
& (Accuracy) & (mIoU) & (F1@0.5) & (AUC) \\ \midrule
GPT-5.6 Sol & \textbf{81.90} & 69.44 & 69.30 & 75.02 \\
Gemini 3.6 Flash & 79.20 & 75.78 & 81.57 & \textbf{75.99} \\
Cosmos3-Super & 72.94 & 76.23 & \textbf{86.97} & 64.66 \\
Qwen3.5-27B & 75.32 & \textbf{76.35} & 85.46 & 24.40 \\
Cosmos-Reason2-8B & 68.60 & 70.26 & 83.88 & 37.69 \\
Qwen3-VL-8B & 68.56 & 73.29 & 59.93 & 33.12 \\
\bottomrule
\end{tabular}
\caption{Single-frame spatial scores against Single Object Tracking, for six representative models. Metrics scaled 0--100.}
\label{tab:spatial-gap}
\end{table}

\subsection{Tracking Anchors and Horizon Analysis}
\label{app:tracking-anchors}

Every system is reported at all three horizons in Table~\ref{tab:tracking-anchors}.

\paragraph{Protocol.} All evaluations are performed on the same underlying object tracklets (200 sequences); the 8-frame, 16-frame, and 32-frame benchmarks differ only in the temporal sampling and clip length. The 8f and 16f datasets share an identical frame-level prefix (stride 15), with the 16f clips extending the 8f sequences by eight additional sampled frames. The 32f benchmark is independently resampled from the same tracklets using a stride of 14, resulting in a longer temporal window rather than a strict continuation of the 16f clips. At the default sampling rate (30 FPS), the three settings correspond to approximately 3.5\,s (8f), 7.5\,s (16f), and 14.5\,s (32f) of video. We additionally report a \emph{freeze rate}: the fraction of predicted boxes that are nearly identical (IoU $\geq 0.95$) to the box predicted on the previous frame.

\paragraph{Baselines.} We evaluate four specialist trackers, a static-box floor, and a random floor on the same 200 tracklets and the exact frames the VLMs see, using the identical Success-AUC scorer. We did not add classical trackers (e.g., KCF, CSRT): on standard benchmarks they are bounded above by the modern trackers and below by these floors, so they would not change where the VLM scores sit.

\begin{table}[htbp]
\centering
\small
\setlength{\tabcolsep}{4pt}

\begin{tabular}{@{}l ccc ccc c@{}}
\toprule
& \multicolumn{3}{c}{\textbf{Success AUC} $\uparrow$} & \multicolumn{3}{c}{\textbf{Freeze rate} $\downarrow$} & \\
\cmidrule(lr){2-4} \cmidrule(lr){5-7}
\textbf{Model} & \textbf{8f} & \textbf{16f} & \textbf{32f} & \textbf{8f} & \textbf{16f} & \textbf{32f} & \textbf{$\Delta$ AUC} \\
\midrule
\multicolumn{8}{@{}l}{\textit{Floors}} \\
Random & 6.55 & 6.27 & 6.12 & 0.00 & 0.00 & 0.00 & $-0.43$ \\
Static box & 24.64 & 18.28 & 13.75 & 99.08 & 99.35 & 99.20 & $-10.89$ \\
\midrule
\multicolumn{8}{@{}l}{\textit{Specialist trackers}} \\
SAM~3 & 79.92 & \textbf{82.25} & \textbf{80.47} & 17.58 & 30.66 & 29.83 & $+0.55$ \\
SAMURAI & \textbf{80.91} & 81.16 & 79.85 & 18.00 & 30.23 & 30.01 & $-1.06$ \\
SUTrack & 79.12 & 81.30 & 78.50 & 16.33 & 28.43 & 27.77 & $-0.62$ \\
MCITrack & 77.74 & 79.44 & 76.61 & 16.67 & 27.53 & 27.29 & $-1.13$ \\
\midrule
\multicolumn{8}{@{}l}{\textit{Vision-language models}} \\
Gemini 3.6 Flash & 75.99 & 70.77 & 66.25 & 23.75 & 36.46 & 36.66 & $-9.74$ \\
GPT-5.6 Sol & 75.02 & 74.75 & 70.91 & 22.75 & 37.32 & 34.96 & $-4.11$ \\
Cosmos3-Super & 64.66 & 61.32 & 45.32 & 32.08 & 48.74 & 58.99 & $-19.34$ \\
Cosmos-Reason2-8B & 37.69 & 30.64 & 17.98 & 77.67 & 84.57 & 92.96 & $-19.71$ \\
Qwen3-VL-8B & 33.12 & 23.27 & 15.67 & 38.42 & 47.20 & 48.50 & $-17.45$ \\
\bottomrule
\end{tabular}
\caption{Tracking anchors and horizon analysis. Success AUC and freeze rate at three horizons, scaled 0--100. The 8-frame column is the Success AUC reported on the public leaderboard.}
\label{tab:tracking-anchors}
\end{table}

\paragraph{Reading the anchors.} The specialist trackers reach approximately 80 Success AUC and hold steady as the horizon grows, establishing the level a purpose-built system reaches on these annotations. Frontier VLMs come within roughly 5 points of them at 8 frames but degrade with horizon (GPT-5.6 Sol $-4.1$, Gemini 3.6 Flash $-9.7$ from 8f to 32f). Open-weight behavior is heterogeneous. Cosmos3-Super remains well above the static floor at 32 frames (45.3 versus 13.8 AUC), although it drops 19.3 points as the horizon increases. Cosmos-Reason2-8B and Qwen3-VL-8B finish much closer to the static floor, at 18.0 and 15.7 versus 13.8. Long-horizon degradation is therefore broadly observed, while convergence toward the static baseline is confined to the weaker open-weight models. The tier ordering---specialists, frontier VLMs, strongest open-weight, then 8B-class---is unchanged across all three horizons.

\paragraph{Reading the freeze rate.} Freeze rate must be interpreted jointly with AUC, since a nearly unchanged box can be correct when the target is stationary; freeze rate alone is not an error rate. The specialist trackers provide an empirical reference that itself rises with horizon: 16--18\% at 8 frames against 27--31\% at 16 and 32 frames. The static-box floor, which by construction never updates, sits at 99\% throughout, so the metric is bracketed at both ends. Cosmos-Reason2-8B has a much higher freeze rate of 93\% together with only 18.0 AUC at 32 frames, consistent with a failure to update its prediction. Qwen3-VL-8B has a lower freeze rate of 49\% but only 15.7 AUC, consistent with continued box updates that drift off target. We therefore use freeze rate to distinguish these two failure modes, without interpreting the specialist rate as the true fraction of stationary targets.

The track covers 200 objects across 102 videos, in line with established tracking test sets (GOT-10k validation: 180 videos; LaSOT: 280). What separates VLMs from purpose-built trackers on this track is persistence over time rather than box regression: at short horizons the frontier models are close to the specialists, and the gap opens as the horizon extends.

\subsection{Domain-Specific Vulnerabilities}

Model performance varies significantly across operational domains, revealing sensitivity to environmental priors (Table~\ref{tab:domain}).

\begin{table}[htbp]
\centering
\small

\begin{tabular}{@{}lcc@{}}
\toprule
\textbf{Domain} & \textbf{VQA (Acc \%)} & \textbf{DVC ($\text{SODA}_c$)} \\ \midrule
Warehouse & 75.73 & 40.61 \\
Smart Spaces & 55.38 & 30.38 \\
Transportation & 71.51 & 8.35 \\ \midrule
Overall & 69.46 & 29.54 \\
\bottomrule
\end{tabular}
\caption{Cosmos3-Super by operational domain, holding model and annotation pipeline fixed.}
\label{tab:domain}
\end{table}

\paragraph{Unstructured Intent.}
Semantic reasoning is weakest in Smart Spaces, at 55.38 VQA accuracy against 75.73 in Warehouse, which we attribute to ambiguous human behaviour and the absence of structured, task-driven interactions.

\paragraph{Egocentric Bias is Temporal, not Semantic.}
Transportation leaves VQA accuracy nearly intact at 71.51 but reduces dense video captioning to 8.35 from 40.61 in Warehouse. The model still recognises what is in the scene; it loses the ability to say when events begin and end. We attribute this to the mismatch between ego-centric training footage and fixed-camera evaluation perspectives, which is most disruptive to the continuous-motion cues temporal grounding depends on. The domain effect is therefore larger than previously reported, but it falls on the temporal pillar.

\subsection{Pillar Scores}
\label{app:pillars}

The main results table reports the eight task scores and the
\textit{Overall} column, but not the intermediate pillar scores from which
\textit{Overall} is computed. Table~\ref{tab:pillars} supplies them. Each
pillar score is the unweighted mean of its constituent task scores, and
\textit{Overall} is the unweighted mean of the four pillar scores, so a
pillar containing one task carries the same weight as a pillar containing
three. This equal weighting is deliberate: it prevents the spatial pillar,
which has the most tasks, from dominating the aggregate. It does mean the
Spatio-Temporal pillar is identical to the Single Object Tracking column of
Table~\ref{tab:main-results}, since tracking is its only task, and that
tracking therefore carries a quarter of the \textit{Overall} score.
Readers who prefer a task-weighted aggregate can compute one from
Table~\ref{tab:main-results}; it would raise the relative weight of the
spatial pillar, where open-weight models are strongest.

\begin{table}[htbp]
\centering
\small
\setlength{\tabcolsep}{6pt}

\adjustbox{max width=\textwidth}{%
\begin{tabular}{@{}ll cccc c@{}}
\toprule
\textbf{Model} & \textbf{Params} & \textbf{Semantic} & \textbf{Spatial} & \textbf{Temporal} & \textbf{Sp.-Temp.} & \textbf{Overall} \\
& & \tiny(2 tasks) & \tiny(3 tasks) & \tiny(2 tasks) & \tiny(1 task) & \\
\midrule
\multicolumn{7}{@{}l}{\textit{Open-weight models}} \\
Qwen3-VL & 2B & 54.29 & 55.22 & 27.55 & 30.38 & 41.86 \\
Qwen3-VL & 8B & 62.91 & 67.26 & 36.98 & 33.12 & 50.07 \\
Qwen3-VL & 32B & 65.67 & 73.56 & 38.10 & 44.19 & 55.38 \\
Qwen3.5 & 9B & 59.20 & 75.26 & 35.20 & 23.52 & 48.30 \\
Qwen3.5 & 27B & 61.65 & \textbf{79.04} & 31.93 & 24.40 & 49.25 \\
Cosmos-Reason2 & 2B & 59.98 & 63.20 & 33.73 & 26.85 & 45.94 \\
Cosmos-Reason2 & 8B & 66.02 & 74.25 & 39.90 & 37.69 & 54.46 \\
Cosmos-Reason2 & 32B & 71.94 & 42.81$^{\dagger}$ & 40.25 & 33.12 & 47.03$^{\dagger}$ \\
Cosmos3-Edge & 4B & 63.02 & 72.02 & 28.07 & 24.65 & 46.94 \\
Cosmos3-Nano & 16B & 68.91 & 74.84 & 39.73 & 59.21 & 60.67 \\
Cosmos3-Super & 64B & 70.37 & 78.71 & 40.72 & 64.66 & 63.62 \\
Gemma-4-E2B & 2B & 36.92 & 27.55 & 9.08 & 11.58 & 21.28 \\
\midrule
\multicolumn{7}{@{}l}{\textit{Proprietary models}} \\
Gemini 3.1 Flash-Lite & --- & 65.91 & 62.71 & 35.13 & 46.68 & 52.61 \\
Gemini 3.1 Pro & --- & 70.01 & 72.14 & 40.62 & 67.88 & 62.66 \\
Gemini 3.5 Flash-Lite & --- & 68.01 & 61.09 & 28.14 & 55.77 & 53.25 \\
Gemini 3.6 Flash & --- & \textbf{79.41} & 78.85 & 43.83 & \textbf{75.99} & \textbf{69.52} \\
GPT-5.6 Sol & --- & 77.11 & 73.55 & \textbf{46.50} & 75.02 & 68.04 \\
\bottomrule
\end{tabular}%
}
\caption{Pillar scores and their aggregate. Each pillar is the unweighted
mean of its tasks; \textit{Overall} is the unweighted mean of the four
pillars. The Spatio-Temporal pillar consists of Single Object Tracking
alone. All values scaled 0--100; bold denotes the best score per column. Markers as in Table~\ref{tab:main-results}.}
\label{tab:pillars}
\end{table}

\subsection{Where the Frontier Advantage Actually Lies}
\label{app:frontier}

The frontier advantage is concentrated in tracking rather than spread across spatial grounding, and this section sets out the evidence for that.

\paragraph{There is no across-the-board proprietary advantage.} Open-weight models lead 2D Object Localization outright (86.97 against 81.57) and lead 2D Referring Expressions by 1.38 points (77.16 against 75.78), and two proprietary models fall below four open-weight models on the overall pillar average (Table~\ref{tab:main-results}). Proprietary access alone does not confer an edge.

\paragraph{The advantage belongs to the two newest releases, and it is mostly uniform.} GPT-5.6 Sol and Gemini 3.6 Flash lead the best open-weight model on six of eight tasks, by a fairly uniform margin of 3.8 to 8.4 points across spatial, semantic, and temporal tasks alike. Three tasks depart from that band. Two are exceptions in the open-weight direction: Cosmos3-Super leads object localization by 5.4 points, and Qwen3.5-9B leads referring expressions by 1.4. Single Object Tracking is the exception in the other direction, where the frontier lead is 11.3 points, roughly double the typical margin.

\paragraph{The distinctive frontier capability is temporal persistence.} The tracking gap is where the frontier separates, not spatial grounding, where open-weight models in fact lead two of the three tasks, and not temporal localization, where the edge is smallest at 3.8 points because no model of either kind exceeds 55.71 mIoU (Section~\ref{sec:scaling}). The horizon analysis in Appendix~\ref{app:tracking-anchors} is consistent with this reading: frontier models degrade least as the tracking horizon extends, while weaker open-weight models converge toward the static-box floor.

\section{De-identification Pipeline}
\label{app:deid}
\label{app:deidentification}

\paragraph{What counts as PII.} We treat as personally identifiable any information that can identify or locate a person: names, recognizable faces, license plates, addresses, and GPS or location data.

\paragraph{Detection and obfuscation.} Faces, license plates, and video locations were obfuscated using a proprietary detector. Detector output was then verified and corrected by human annotators, so that the released media reflect a human-in-the-loop pass rather than automated obfuscation alone.

\paragraph{Licensing safeguards.} Beyond obfuscation, the NVIDIA Evaluation Data License prohibits biometric identification and demographic profiling, restricting the released data to evaluation and benchmarking use.

\paragraph{Consent.} 70\% of footage was provided by vendors who obtained explicit informed consent from recorded individuals for research and redistribution. A smaller portion was captured in limited spaces with posted notice of camera operation, where individuals incidentally captured consented to fixed-camera monitoring as a condition of site access.

\section{Reproducibility and Experimental Configuration}
\label{app:experimental-details}

\subsection{Compute Resources and Frameworks}
Evaluations for open-weight models (Cosmos-Reason2, Qwen3-VL, Qwen3.5, Cosmos3, Gemma-4) were conducted on the Palmetto 2 cluster at Clemson University, using 7 nodes of 8$\times$ NVIDIA H100 (80GB) GPUs, as well as additional cloud compute instances. Proprietary models (GPT-5.6 Sol and the Gemini family) were accessed via public APIs. We build our evaluation harness on top of VLMEvalKit \cite{vlmevalkit}, extending it with custom metric scripts for automated SODA$_c$ computation, temporal IoU evaluation, and spatial trajectory parsing.

\subsection{Task-Specific Inference Configurations}
\label{app:inference}
All evaluations used greedy decoding (temperature = 0) with 
chain-of-thought disabled (\texttt{enable\_thinking = False}). 
The sampling and pixel budgets used to manage visual granularity and
token limits are given in Table~\ref{tab:inference-configs}. FPS 
ranges reflect model-specific sampling rates.

\begin{table}[htbp]
\centering
\small

\begin{tabular}{@{}llccl@{}}
\toprule
\textbf{Pillar} & \textbf{Task} & \textbf{FPS} & 
\textbf{Max Frames} & \textbf{Max Total Pixels} \\
\midrule
Semantic & EV / VQA & 1.0 -- 4.0  & 256 & $16.7 \times 10^6$ \\
Spatial  & Pointing / RefEx & ---      & 1   & $1.5 \times 10^6$ \\
Temporal & Temporal Localization & 2.0 -- 10.0 & 256 & $16.7 \times 10^6$ \\
Temporal & DVC              & 1.0 -- 4.0  & 128 & $8.4 \times 10^6$ \\
Spatio-Temp. & SOT & --- & 8 / 16 / 32 & $0.9 \times 10^6$ \\
Spatial  & 2D Object Localization & --- & 1 & $1.5 \times 10^6$ \\
\bottomrule
\end{tabular}
\caption{VANTAGE-Bench Task Inference Parameters.}
\label{tab:inference-configs}
\end{table}

\subsection{Code, Data Access, and Release Strategy}
\label{app:release}
To support reproducibility, we provide the VANTAGE-Bench dataset at \href{https://huggingface.co/datasets/nvidia/PhysicalAI-VANTAGE-Bench}{huggingface.co/datasets/nvidia/PhysicalAI-VANTAGE-Bench}
the evaluation code at \href{https://github.com/Clemson-Capstone/VANTAGE-Bench}{github.com/Clemson-Capstone/VANTAGE-Bench}, and the public leaderboard at \href{https://huggingface.co/spaces/clemson-computing/VANTAGE-Bench-Leaderboard}{huggingface.co/spaces/clemson-computing/VANTAGE-Bench-Leaderboard}. Project information is collected at \href{https://vantage-bench.org/}{vantage-bench.org}. To preserve long-term integrity against training-data contamination, we adopt a held-out evaluation protocol. We publicly release (i) all input media, (ii) all prompts and questions, and (iii) the evaluation harness. Ground-truth annotations are held-out to be used by the official leaderboard server. The leaderboard is live and open for submissions, which are uploaded as JSON in LLaVA format.

\section{Reported Uncertainty}
\label{app:uncertainty}

We report two kinds of variability, which answer different questions: how
precisely a score is estimated from the items in the test set, and how
reproducible it is across runs.

\subsection{Item-Level Confidence Intervals}
\label{app:bootstrap}

Because all evaluation uses greedy decoding at temperature 0, run-to-run
variance is small (Section~\ref{app:stability}, below); the dominant source of
uncertainty is which items happen to be in the test set. We therefore
report 95\% percentile bootstrap confidence intervals over test items,
resampling items with replacement 10{,}000 times per model--task cell and
recomputing the metric on each resample. For Single Object Tracking the
resampling unit is the object tracklet ($n=200$) rather than the
individual frame, since frames within a tracklet are strongly correlated.

The half-width of each interval is given in Table~\ref{tab:ci}. Interval width
is governed by the number of scored items, so it varies far more across
tasks than across models. Referring expressions and object localization
are the tightest (median $\pm$1.00 and $\pm$1.27 respectively), while event
verification is the widest by a wide margin (median $\pm$7.47), reflecting
its smaller item pool and binary scoring. The practical consequence is
that the comparisons we draw in Section~\ref{sec:results} should be read
against very different resolutions: a two-point difference is meaningful
on referring expressions and is not meaningful on event verification.

\begin{table}[htbp]
\centering
\scriptsize
\setlength{\tabcolsep}{3pt}

\adjustbox{max width=\textwidth}{%
\begin{tabular}{@{}ll cc ccc cc c@{}}
\toprule
& & \multicolumn{2}{c}{\textit{Semantic}} & \multicolumn{3}{c}{\textit{Spatial}} & \multicolumn{2}{c}{\textit{Temporal}} & \textit{Sp.-Temp.} \\
\cmidrule(lr){3-4} \cmidrule(lr){5-7} \cmidrule(lr){8-9} \cmidrule(lr){10-10}
\textbf{Model} & \textbf{Params} & \textbf{VQA} & \textbf{EV} & \textbf{Point.} & \textbf{RefEx} & \textbf{Obj Loc.} & \textbf{Temp Loc.} & \textbf{DVC} & \textbf{SOT} \\
\midrule
\multicolumn{10}{@{}l}{\textit{Open-weight models}} \\
Qwen3-VL & 2B & 63.85 {\tiny $\pm$2.76} & 44.73 {\tiny $\pm$7.56} & 53.13 {\tiny $\pm$3.08} & 65.12 {\tiny $\pm$1.10} & 47.42 {\tiny $\pm$2.68} & 35.21 {\tiny $\pm$2.23} & 19.90 {\tiny $\pm$1.77}  & 30.38 {\tiny $\pm$2.83} \\
Qwen3-VL & 8B & 66.44 {\tiny $\pm$2.68} & 59.39 {\tiny $\pm$7.64} & 68.56 {\tiny $\pm$2.94} & 73.29 {\tiny $\pm$1.01} & 59.93 {\tiny $\pm$3.11} & 44.32 {\tiny $\pm$2.49} & 29.64 {\tiny $\pm$2.26} & 33.12 {\tiny $\pm$2.90} \\
Qwen3-VL & 32B & 71.30 {\tiny $\pm$2.55} & 60.04 {\tiny $\pm$7.66} & 75.62 {\tiny $\pm$2.64} & 72.39 {\tiny $\pm$1.00} & 72.67 {\tiny $\pm$2.51} & 46.80 {\tiny $\pm$2.40} & 29.40 {\tiny $\pm$2.87} & 44.19 {\tiny $\pm$2.98} \\
Qwen3.5 & 9B & 71.63 {\tiny $\pm$2.51} & 46.78 {\tiny $\pm$7.61} & 71.74 {\tiny $\pm$2.79} & 77.16 {\tiny $\pm$0.90} & 76.87 {\tiny $\pm$1.81} & 42.53 {\tiny $\pm$2.49} & 27.88 {\tiny $\pm$2.87} & 23.52 {\tiny $\pm$2.47} \\
Qwen3.5 & 27B & 67.95 {\tiny $\pm$2.68} & 55.34 {\tiny $\pm$7.73} & 75.32 {\tiny $\pm$2.64} & 76.35 {\tiny $\pm$0.89} & 85.46 {\tiny $\pm$0.64} & 36.88 {\tiny $\pm$2.22} & 26.97 {\tiny $\pm$2.53} & 24.40 {\tiny $\pm$2.35} \\
Cosmos-Reason2 & 2B & 64.69 {\tiny $\pm$2.72} & 55.27 {\tiny $\pm$7.97} & 59.70 {\tiny $\pm$3.08} & 56.33 {\tiny $\pm$1.24} & 73.56 {\tiny $\pm$1.62} & 38.95 {\tiny $\pm$2.53} & 28.52 {\tiny $\pm$3.12} & 26.85 {\tiny $\pm$2.71} \\
Cosmos-Reason2 & 8B & 67.95 {\tiny $\pm$2.72} & 64.09 {\tiny $\pm$7.70} & 68.60 {\tiny $\pm$2.89} & 70.26 {\tiny $\pm$1.02} & 83.88 {\tiny $\pm$0.89} & 47.30 {\tiny $\pm$2.53} & 32.50 {\tiny $\pm$3.50} & 37.69 {\tiny $\pm$3.12} \\
Cosmos-Reason2 & 32B & 70.29 {\tiny $\pm$2.59} & 73.58 {\tiny $\pm$7.00} & 74.03 {\tiny $\pm$2.74} & 50.04 {\tiny $\pm$1.05} & 4.35$^{\dagger}$ {\tiny $\pm$0.91} & 50.47 {\tiny $\pm$2.65} & 30.04 {\tiny $\pm$3.18} & 33.12 {\tiny $\pm$3.02} \\
Cosmos3-Edge & 4B & 62.68 {\tiny $\pm$2.76} & 63.36 {\tiny $\pm$7.51} & 67.76 {\tiny $\pm$2.84} & 69.27 {\tiny $\pm$1.09} & 79.02 {\tiny $\pm$0.70} & 39.51 {\tiny $\pm$2.57} & 16.64 {\tiny $\pm$2.85} & 24.65 {\tiny $\pm$2.81} \\
Cosmos3-Nano & 16B & 68.95 {\tiny $\pm$2.68} & 68.88 {\tiny $\pm$7.16} & 74.83 {\tiny $\pm$2.69} & 75.57 {\tiny $\pm$0.99} & 74.11 {\tiny $\pm$1.74} & 48.04 {\tiny $\pm$2.61} & 31.42 {\tiny $\pm$3.81} & 59.21 {\tiny $\pm$3.75} \\
Cosmos3-Super & 64B & 69.46 {\tiny $\pm$2.59} & 71.28 {\tiny $\pm$7.12} & 72.94 {\tiny $\pm$2.74} & 76.23 {\tiny $\pm$0.98} & 86.97 {\tiny $\pm$0.70} & 51.90 {\tiny $\pm$2.57} & 29.54 {\tiny $\pm$3.31} & 64.66 {\tiny $\pm$3.67} \\
Gemma-4-E2B & 2B & 46.19 {\tiny $\pm$2.85} & 27.65 {\tiny $\pm$4.52} & 31.74 {\tiny $\pm$2.94} & 5.52 {\tiny $\pm$0.40} & 45.40 {\tiny $\pm$0.74} & 9.02 {\tiny $\pm$1.10} & 9.14 {\tiny $\pm$1.83} & 11.58 {\tiny $\pm$1.31} \\
\midrule
\multicolumn{10}{@{}l}{\textit{Proprietary models}} \\
Gemini 3.1 Flash-Lite & --- & 68.03 {\tiny $\pm$2.72} & 63.78 {\tiny $\pm$7.47} & 63.68 {\tiny $\pm$2.99} & 52.70 {\tiny $\pm$1.41} & 71.74 {\tiny $\pm$1.34} & 37.72 {\tiny $\pm$2.36} & 32.54 {\tiny $\pm$2.46} & 46.68 {\tiny $\pm$3.89} \\
Gemini 3.1 Pro & --- & 71.46 {\tiny $\pm$2.55} & 68.57 {\tiny $\pm$7.23} & 72.04 {\tiny $\pm$2.79} & 67.17 {\tiny $\pm$1.27} & 77.21 {\tiny $\pm$1.73} & 45.69 {\tiny $\pm$2.36} & 35.55 {\tiny $\pm$2.82} & 67.88 {\tiny $\pm$3.98} \\
Gemini 3.5 Flash-Lite & --- & 65.61 {\tiny $\pm$2.68} & 70.41 {\tiny $\pm$7.06} & 43.08 {\tiny $\pm$3.03} & 74.46 {\tiny $\pm$0.97} & 65.73 {\tiny $\pm$1.27} & 24.00 {\tiny $\pm$2.67} & 32.29 {\tiny $\pm$2.39} & 55.77 {\tiny $\pm$4.56} \\
Gemini 3.6 Flash & --- & 76.82 {\tiny $\pm$2.38} & 82.00 {\tiny $\pm$6.26} & 79.20 {\tiny $\pm$2.54} & 75.78 {\tiny $\pm$0.87} & 81.57 {\tiny $\pm$0.61} & 51.50 {\tiny $\pm$2.67} & 36.15 {\tiny $\pm$3.16} & 75.99 {\tiny $\pm$3.44} \\
GPT-5.6 Sol & --- & 78.08 {\tiny $\pm$2.38} & 76.14 {\tiny $\pm$6.72} & 81.90 {\tiny $\pm$2.39} & 69.44 {\tiny $\pm$0.91} & 69.30 {\tiny $\pm$0.72} & 55.71 {\tiny $\pm$2.65} & 37.28 {\tiny $\pm$2.88} & 75.02 {\tiny $\pm$3.34} \\
\bottomrule
\end{tabular}%
}
\caption{Scores with 95\% item-level bootstrap confidence-interval half-widths ($\pm$), for
every cell of Table~\ref{tab:main-results}, in points on the 0--100 scale
(10{,}000 resamples; tracking resampled at the tracklet level) \cite{miller2024errorbars}. Scored
items per task: VQA 1195; EV 163; Point. 1005; RefEx 3276; Obj Loc. 628; Temp Loc. 203; DVC 104; SOT 200.}
\label{tab:ci}
\end{table}

\subsection{Seed Variance}
\label{app:stability}

To confirm that reported differences are not seed-driven, we ran Qwen3-VL-8B three times with different random seeds (Table~\ref{tab:stability}). Because decoding is greedy at temperature 0, this variance is expected to be small, and it is. All five tasks measured show a standard deviation below 0.5 points, with 2D Spatial Pointing and Single Object Tracking the largest at 0.42 and 0.40. This establishes that individual cells are reproducible across runs. It is not a basis for judging whether two models differ: seed variance holds the item set fixed and therefore omits the dominant source of uncertainty in benchmark scores, which is item sampling \cite{miller2024errorbars}. For model comparisons, the item-level intervals above are the relevant quantity.

The dominant source of uncertainty is item sampling, which seed runs cannot measure. The variance reported here should therefore not be read as a confidence interval on the scores themselves; item-level bootstrap intervals for every model and task are reported separately in Appendix~\ref{app:bootstrap}.

\begin{table}[H]
\centering
\small

\begin{tabular}{@{}llc@{}}
\toprule
\textbf{Pillar} & \textbf{Task (metric)} & \textbf{Std. dev. across 3 seeds} \\
\midrule
Semantic & Event Verification (Macro F1) & $0.31$ \\
Semantic & Video Question Answering (Accuracy) & $0.09$ \\
Spatial & 2D Spatial Pointing (Accuracy) & $0.42$ \\
Temporal & Dense Video Captioning (SODA$_c$) & $0.32$ \\
Spatio-Temp. & Single Object Tracking (AUC) & $0.40$ \\
\bottomrule
\end{tabular}
\caption{Seed variance on VANTAGE-Bench (Qwen3-VL-8B), standard deviation across three random seeds, scaled 0--100. This reflects decoding randomness only, not item sampling.}
\label{tab:stability}
\end{table}

\end{document}